\documentclass[journal]{IEEEtran}
\usepackage[T1]{fontenc} 
\usepackage{amsmath}
\usepackage{bm} 
\usepackage{cite}

\usepackage[caption=false,farskip=0pt,labelfont={bf}]{subfig}
\usepackage{graphicx}
\usepackage[colorlinks,linkcolor=blue]{hyperref}
\usepackage{bbding}
\usepackage{booktabs}
\usepackage{tikz,xcolor}

\usepackage[T1]{fontenc} 
\usepackage{amsmath}
\usepackage{bm} 
\usepackage{cite}

\usepackage{graphicx}
\usepackage[colorlinks,linkcolor=blue]{hyperref}
\usepackage{bbding}
\usepackage{booktabs}
\usepackage{tikz,xcolor}

\usepackage{amssymb, amsmath, amsthm, bm}
\usepackage[noend]{algpseudocode}
\usepackage{algorithmicx,algorithm}
\usepackage[caption=false]{subfig}
\usepackage{epstopdf}
\usepackage{diagbox}
\usepackage{listings}
\usepackage{multirow}
\definecolor{lime}{HTML}{A6CE39}
\DeclareRobustCommand{\orcidicon}{%
	\begin{tikzpicture}
	\draw[lime, fill=lime] (0,0)
	circle [radius=0.16]
	node[white] {{\fontfamily{qag}\selectfont \tiny ID}};    \draw[white, fill=white] (-0.0625,0.095)
	circle [radius=0.007];    \end{tikzpicture}
	\hspace{-2mm}}
\foreach \x in {A, ..., Z}{%
	\expandafter\xdef\csname orcid\x\endcsname{\noexpand\href{https://orcid.org/\csname orcidauthor\x\endcsname}{\noexpand\orcidicon}}}

\newtheorem{rmrk}{Remark}

\begin{document}
	
\title{Improve Deep Image Inpainting by Emphasizing the Complexity of Missing Regions}

\author{Yufeng~Wang\orcidA,
        Dan~Li,
        Cong~Xu\orcidB,
        Min~Yang\orcidC
\thanks{This work was supported in part by the National Natural Science Foundation of China under grant 11771257. (\emph{Corresponding author: Min Yang.})}
\thanks{
	Yufeng Wang, Dan Li, Cong Xu and Min Yang are with the School of Mathematics and Information Sciences,
	Yantai University, Yantai 264005, China (email: ytuyufengwang@163.com; danliai@hotmail.com; congxueric@gmail.com; yang@ytu.edu.cn)
}
}%

\maketitle

\begin{abstract}
Deep image inpainting research mainly focuses on constructing various neural network architectures or imposing novel optimization objectives.
However, on the one hand, building a state-of-the-art deep inpainting model is an extremely complex task,
and on the other hand, the resulting performance gains are sometimes very limited.
We believe that besides the frameworks of inpainting models,
lightweight traditional image processing techniques, which are often overlooked, can actually be helpful to these deep models.
In this paper, we enhance the deep image inpainting models with the help of classical image complexity metrics.
A knowledge-assisted index composed of missingness complexity and forward loss is presented to guide the batch selection in the training procedure.
This index helps find samples that are more conducive to optimization in each iteration and ultimately boost the overall inpainting performance.
The proposed approach is simple  and can be plugged into many deep inpainting models by changing only a few lines of code.
We experimentally demonstrate the improvements for several recently developed image inpainting models on various datasets.
\end{abstract}


%

\section{Introduction}

\IEEEPARstart{I}{mage} inpainting aims to restore missing regions of corrupted images with realistic content,
and has a wide range of applications in photo editing,
de-captioning and other scenarios where people might want to remove unwanted objects from their photos \cite{Jam2021, Zhou2021, Shetty2018, Song2019}.
Recent image inpainting models usually rely on complicated neural networks and well-designed loss functions to produce satisfactory results \cite{Nazeri2019, Yan2018}.
However, with the development of research,
the inpainting models have become more and more complicated, while the improvements sometimes become limited.
Therefore, it is a meaningful direction to develop general ``plug and play'' techniques  that can help existing image inpainting  models to enhance their performance further.

It is well known that batch samples have a significant impact on the final performance of deep learning models \cite{Chang2017, Johnson2018,Katharopoulos2018,Mindermann2021}.
In recent years,
there has been a surge of research in developing sample selection methods in various learning tasks,
e.g. image classification \cite{Johnson2018}, object detection \cite{Lin2017,Shrivastava2016} and reinforcement learning \cite{Schaul2016}.
These works mainly use the training losses \cite{Jiang2019, Kawaguchi2020} to estimate the priority of each sample,
and then select those that are most conducive to the learning to update neural networks.
However, we found that this simple selection technique could not be applied for image inpainting models,
because it will cause only those samples with high--missingness complexity to be selected,
and the samples of low-missingness complexity may never be used in training.
Such learning bias can not improve the inpainting models rather than deteriorate their performance (see Section \ref{RTM} for detailed analysis and experimental results in Section \ref{Sebias}).

In this paper,
inspired by the correlation of training loss and missingness complexty,
we incorporate the complexity of missing regions into batch selection
and present a novel knowledge-assisted sample selection method that are suitable for image inpainting tasks.
Specifically,
at each training iteration,
the forward loss and the missingness complexity of each sample are first calculated.
Then a selection index mainly determined by the ratio of loss and complexity is presented to guide the batch selection.
It is worth to point out the missingness complexity has nothing to do with the neural networks, and
can be computed via straightforward mathematical calculations.
It is also noteworthy that our approach is easy to implement and does not need to change existing neural networks or loss functions of inpainting models.
The presented method gives a good consideration to intuitive explanation of the problem  and cognitive capacity of deep learning.
A little extra code is enough to promote the final inpainting quality.

The main contributions of the paper are summarized as follows:
\begin{itemize}
    \item
    We improve the quality of image inpainting from the perspective of batch selection.
    The developed selection method breaks the dilemma of the current selection methods in the application to image inpainting,
    and can be used as a ``plug and play''  technique to improve existing deep image inpainting models.
    \item
    Compared to purely loss-based selection methods,
    our selection indicator combines deep learning with traditional image metrics,
    and incorporates missingness complexity to avoid batch selection bias in training.
    Such a combination demonstrates  the power of promoting deep learning with classical image analysis tools.
    \item
    The experimental results show that our approach is efficient and can be applied in many recently developed image inpainting models
    to enhance their performance  by changing only a few lines of code.
\end{itemize}

\section{Related Works}

\subsection{Image inpainting}
Traditional image inpainting methods fill the corrupted regions by diffusion-based methods \cite{Ballester2001, Li2017}
or patch-based methods \cite{Huang2014, Ghorai2019, Ding2019}.
In recent years, due to its powerful semantic representation ability,
deep learning has been widely used in image inpainting and achieved superior performance \cite{Pathak2016, Yu2018, Liu2018A, Wang2018, Zeng2019, Li2020, Yu2020}.
ContextEncoder \cite{Pathak2016}, one of pioneer learning-based methods,
proposed an encoder-decoder structure to complete the task of image inpainting.
ContextualAttention \cite{Yu2018} introduced additional context attention layers to bring global features  for subsequent networks.
In order to handle irregular corrupted images, PC \cite{Liu2018A} proposed to use partial convolutions and considered an automatic mask update step.
PEN-Net \cite{Zeng2019} adopted  a U-Net architecture and restored the missing part by encoding the context semantics of the input and then decoding the learned features back to the image.
RFR \cite{Li2020} is a progressive inpainting model that simulates the way that humans solve puzzles and  gradually inpaints incomplete images.
RN \cite{Yu2020} divided the spatial pixels into different regions according to the input masks,
and took the mean and variance of each region for normalization.

In spite of the promising advances brought by these deep inpainting models,
the corresponding learning frameworks has become more and more complicated.
On the contrary, the performance gains sometimes seem limited.
Therefore, it is interesting to develop some simple but general "plug and play" approaches that can be applied to further improve existing inpainting models.

\subsection{Sample selection methods}
Many literatures have found that  well-chosen mini-batch samples can help promote the performance of deep learning models. \cite{Chang2017,Fan2017, Jiang2019, Katharopoulos2018,Kawaguchi2020,Song2020}.
Fan et al. \cite{Fan2017} proposed a selection technique that prioritizes the samples with high forward losses of the training dataset to improve binary classification and regression.
Compared to \cite{Fan2017},
Kawaguchi et al. \cite{Kawaguchi2020} restricted  the selection in a random subset of the entire dataset at each epoch.
Jiang et al. \cite{Jiang2019} generated a sample distribution at each epoch using the training losses of all samples,
then probabilistically skipping the backward pass for samples exhibiting low loss.
Different from the above selection methods that determine batch samples using the current forward losses,
some works considered the variation of losses for each sample during training period.
Chang et al. \cite{Chang2017} maintained the historical losses of each example,
and took those samples  with larger loss--variances into the mini-batch.
Song et al.\cite{Song2020} introduced additional sliding windows,
so that the uncertainty of each sample can be adaptively adjusted along with the training.

These selection methods mainly rely on training losses to find bath samples that are beneficial for learning.
However, we found that they are not applicable for image inpainting due to the following two reasons.
First, in an image inpainting task,
the training loss of any sample is closely correlated to its missingness complexity,
i.e., high-loss samples are usually accompanied by high-missingness complexity.
Simply using training losses to select batch samples will make samples with low-minsingness complexity samples rarely used in training.
Such a learning bias may alleviate the performance of the inpainting model instead of improving it.
Secondly, for the same sample, its mask often changes with the iteration,
so the variation of its historical losses is not as informative as in other learning tasks.

\subsection{Image complexity}
Image complexity is a kind of classical image measurement tool,
which has been successfully used in a variety of computer vision tasks,
e.g., image restoration \cite{Bertalmio2003,Perkio2009},
image classification \cite{Romero2012}, and target automatic extraction \cite{Song2014}.
However, the definition of the complexity of an image  is not as straightforward as it seems.
Different complexity metrics could explain the complexity of the image from different aspects.
For example, the Gray-Level Co-occurrence Matrix (GLCM) \cite{Haralick1973}  has been widely used  for texture analysis.
The statistical properties of GLCM, e.g. second moments, correlation coefficients or entropy,
can be used to express the texture information of images \cite{Song2014, Aouat2021}.
Spatial Information (SI) \cite{Yu2013} is a compression-based image complexity metric that takes edge energy to obtain the spatial information matrix of the image,
and expresses the complexity of the image through the matrix's mean, variance, and root mean square.
Total Variation (TV) treats the image as a function with limited variation \cite{Aujol2009}
and uses a two or multi-dimensional variations mechanisms to evaluate the texture complexity of the image,
which has been proven effective in many image tasks \cite{Chamb2012, Chochia2015}.

These traditional complexity metrics have good interpretability
and can be implemented by simple mathematical formulas without the need of tedious learning procedures.

\section{Incorporating Missingness Complexity into Sample Selection }
Let $  \mathcal{D}  \subseteq \mathbb{R}^{C \times H \times W} $ denote an image dataset.
For each $ \mathbf{x} \in  \mathcal{D}  $, there exists a corresponding binary mask  $ \mathbf{m}= \{0,1\}^{H \times W} $,
where $ m^{i,j} = 1 $ means that the pixels of all channels at $(i,j)$ are observed, and $ m^{i,j} = 0 $ means that the corresponding pixels are missing.
An inpainting model tends to use the observed part  $  \mathbf{x}\odot  \mathbf{m} $ to reconstruct the missing region $ \mathbf{x}\odot  (\mathbf{1}- \mathbf{m}) $.

In the following, we shall first recall some sample selection methods and introduce the complexity metrics to be used in the paper.
Then the relationship between the training loss and the complexity of the corrupted region is to be provided.
According to this relationship, it can be found that simply using training losses to select mini-batch samples could lead to  a severe selection bias.
Finally, a novel selection metric is proposed to remedy this bias and help improve deep image inpainting models.

\subsection{Loss-based selection methods}

A common loss-based sample selection index can be expressed as
\begin{align}
    \mathcal{S}(\mathbf{x}) = f(\mathcal{L}(\mathbf{x})),
\label{score}
\end{align}
where $\mathcal{L}( \cdot )$ denotes the training loss of the sample $\mathbf{x}$ and $f( \cdot )$ is a specific selection function.

In \cite{Fan2017},
Fan et al.  proposed to select top-$ k $ samples with the largest individual loss in the dataset at each iteration,
and the corresponding selection index can be defined as
\begin{align}
       \mathcal{S}(\mathbf{x}) = \mathbf{argmax}(\mathcal{L}_{top-k}(\mathbf{x})), \mathbf{x} \in \mathcal{D}.
\label{Fan}
\end{align}
Kawaguchi et al. \cite{Kawaguchi2020} restricted the selection range to a random subset $\mathcal{B}$ of the entire dataset,
i.e.,
\begin{align}
       \mathcal{S}(\mathbf{x}) = \mathbf{argmax}(\mathcal{L}_{top-k}(\mathbf{x})), \mathbf{x} \in \mathcal{B} \subset \mathcal{D}.
\label{Kawaguchi}
\end{align}
Jiang et al. \cite{Jiang2019} calculated the selection probability of each sample as follows
\begin{align}
\mathcal{S}(\mathbf{x}) = [\mathbf{CDF_R}(\mathcal{L}(\mathbf{x}))]^{\beta}, \mathbf{x} \in \mathcal{D},
\label{Jiang}
\end{align}
where $\mathbf{CDF_R}( \cdot )$ is a cumulative distribution function (CDF)  and $\beta $ is a positive  constant to adjust the deviation of those large--loss samples.

\subsection{Missingness complexity measures}

Spatial information (SI) \cite{Yu2013} is an indicator of edge energy
and has been commonly used as the basis for estimating image complexity.
At each pixel in the corrupted region,
let $ s_h $ and $ s_v $ be gray-scale  images filtered with horizontal and vertical Sobel kernels \cite{Kazakova2004}, respectively.
Then $  = \sqrt{s_h^2 + s_v^2} $ represents the spatial information level of the corresponding  pixel.
The root-mean-square of  the $SI_r $ values of all pixels in the missing region can reflect its complexity.
That is
\begin{align}
    SI = \sqrt{\frac{1}{N_m}\sum SI_r^2},
\label{sirms_eq}
\end{align}
where $ N_m $ denotes the number of the pixels in the missing region.

Next, we define the Gray-Level Co-occurrence Matrix (GLCM).
For an arbitrary image with 256 distinct gray levels,
denote by $ N_{i,j} $ the number of the pixel pairs that have gray levels $ i, j $ and whose position are
specified by a position operator. The
position operator describes two parameters for the pixel
pairs: pixel pairs spacing  and pixel pairs direction \cite{Aouat2021}.
Then the entries of a GLCM satisfy
$   G(i,j)=N_{i,j}/\sum_{i,j=0}^{255} N_{i,j} $, $ 0 \leq i, j \leq 255 $.
One can use $ G $ to estimate the texture entropy of the corrupted region.
That is
\begin{align}
     EG =-\sum_{i,j=0}^{255}  G(i,j) \ln G(i,j)
\label{glcm_eq}
\end{align}

Total Variation (TV) measures the complexity of an image in terms of the variation between adjacent pixels.
The following metric  can be used to evaluate the missingness complexity:
\begin{align}
\begin{split}
TV =& \sum_{i=1}^{H} \sum_{j=1}^{W-1} |g^{i,j+1} - g^{i,j}| (1-\mathbf{m}^{i,j})
   \\[5pt]
    +&\sum_{i=1}^{H-1} \sum_{j=1}^{W} |g^{i+1,j} -g^{i,j}|  (1-\mathbf{m}^{i,j}).
\end{split}
\label{tv_eq}
\end{align}
where $ g $ denotes the gray value of the corresponding pixel and $\mathbf{m} $ is the mask of the image.

\begin{figure}[hbt]
 \centering
 \scalebox{0.5}{\includegraphics{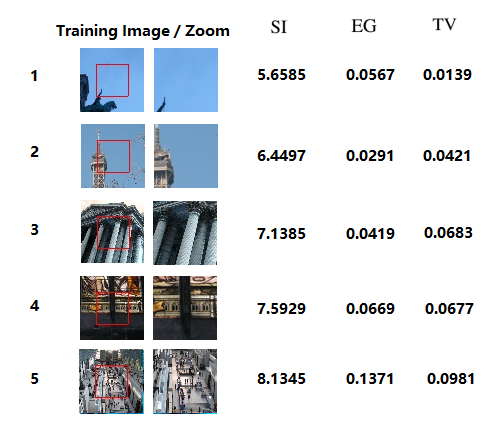}}
 \caption{An illustration of various missingness complexity metrics. The red rectangle represents the missing region.}
 \label{complexity_visual}
\end{figure}

Different metrics can reflect the complexity of the image from different perspectives.
An intuitive demonstration of missingness complexity (before normalization) is illustrated in Fig. \ref{complexity_visual}.

\subsection{Relationship between missingness complexity and training loss}
\label{RTM}

\begin{figure}[hbt]
 \centering
 \scalebox{0.35}{\includegraphics{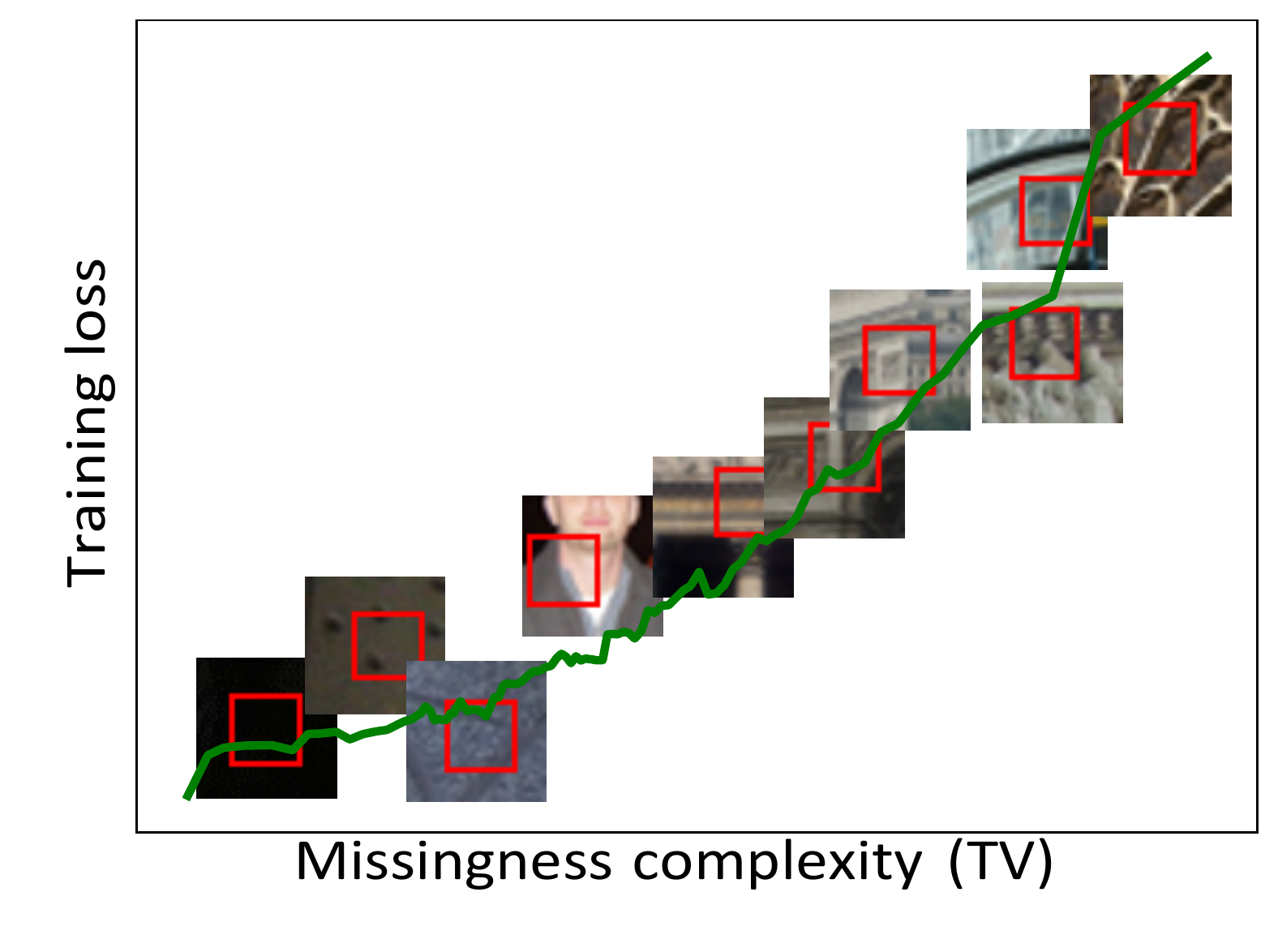}}
 \caption{The change curve of the correlation between missingness complexity and training loss.
              The red rectangle represents the missing region.}
 \label{loss_tv_correlation}
\end{figure}

Now we  are to investigate the relationship between the missingenss complexity and the training loss of each sample.
To this end, we take RFR\cite{Li2020} as an example and use TV \eqref{tv_eq} as the complexity measure.
We conduct the model on the CelebA \cite{Liu2018B} dataset for 5 epochs,
and then calculate the training loss and the missiningness complexity for each sample.
It is found that those larger loss samples are usually accompanied by higher missingness complexity (See Fig. \ref{loss_tv_correlation}).
This is not an accidental phenomenon,
and similar results could be observed for other inpainting models and complexity metrics.

\begin{figure}[htb]
\center
\subfloat[loss-based method]{\label{fig:mdleft}{\includegraphics[width=0.23\textwidth]{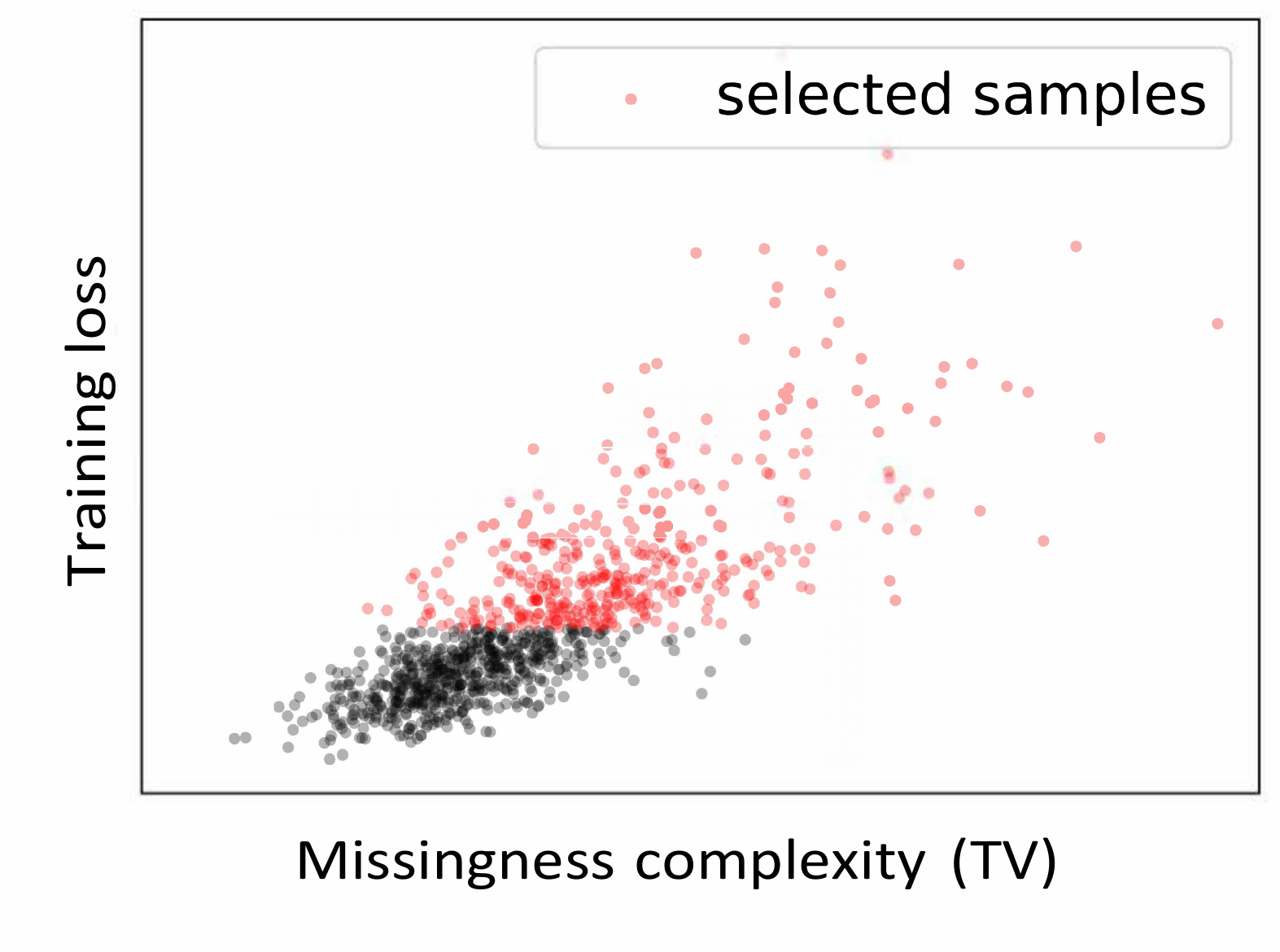}}}
\subfloat[our method]{\label{fig:mdleft}{\includegraphics[width=0.23\textwidth]{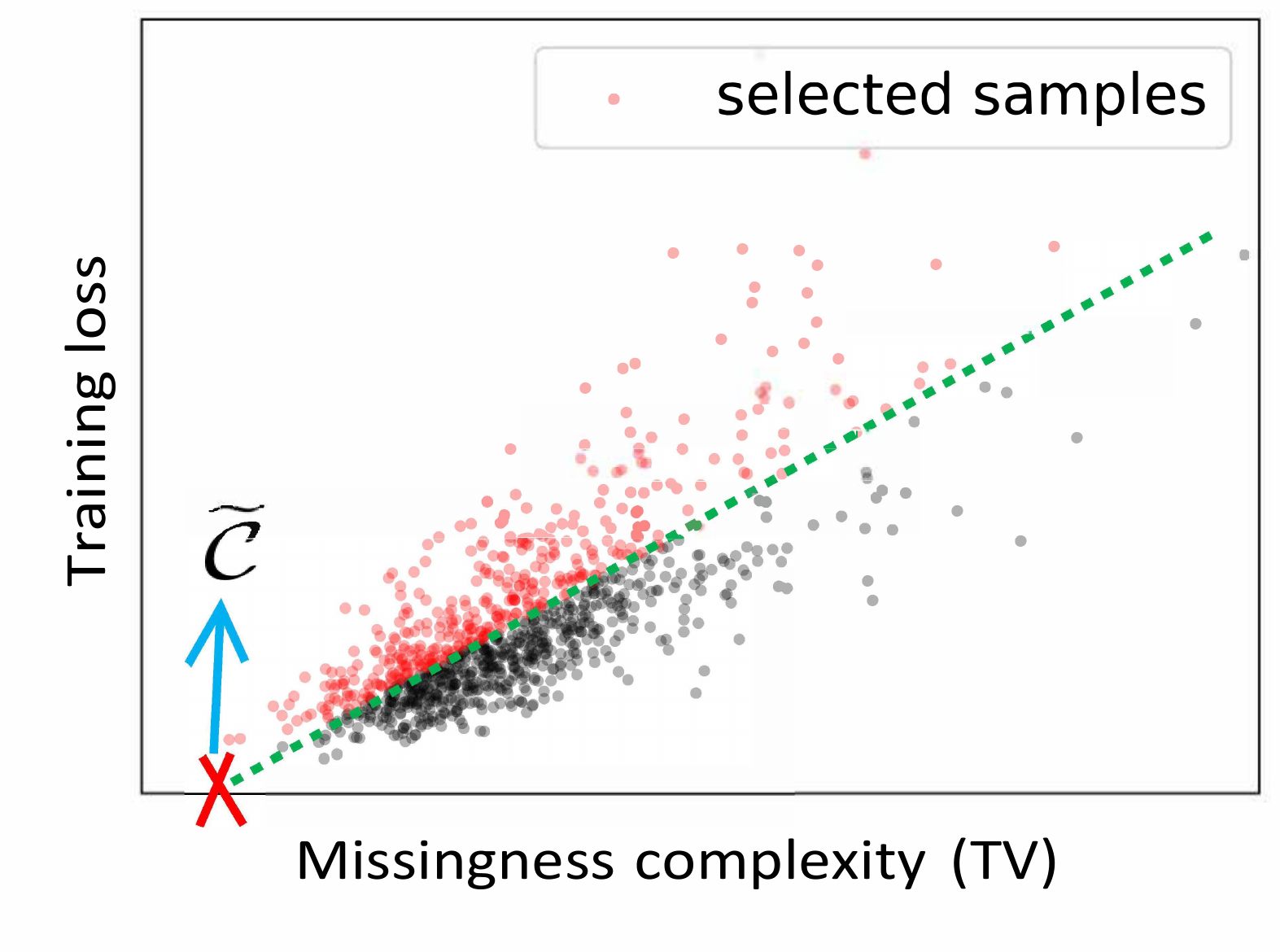}}}
\caption{ The batch samples selected by previous loss-based  method and our method.}
\label{select}
\end{figure}

Therefore, if one expects to use the common loss-based sample selection method to improve the performance of the inpainting model,
he will fall into a trap.
As shown in Fig. \ref{select} (a),
only those samples with high-missingness complexity are to be used in training,
while low-missingness complexity ones almost have no chance to participate in training.
However, low-missingness complexity samples are beneficial to the model to find the optimal solution faster, especially in the early training stage (see experimental results in Section \ref{Time}).
Thus, neglecting these simple samples can  degrade model's performance rather than improve it.

\subsection{The proposed selection criteria}
As discussed in the previous section,
the common loss-based sample selection methods may lead to unsatisfactory  selection  bias for deep inpainting models.
In this section, we aim to remedy such bias and
propose a novel selection criterion incorporating the missingness complexity.

According to the correlation between training loss and missingness complexity,
we argue that an ideal sample selection result should be similar to Fig. \ref{select} (b),
i.e.,
under the condition of covering the complexity distribution, the samples with large losses  are preferred.

Thus, we propose the new selection criterion as follows

\begin{align}
\label{select_indicate}
S(\mathbf{x}, \mathbf{m}) =
f\left(\frac{\mathcal{L}(\mathbf{x}, \mathbf{m})}{\max\left(|\mathcal{C}(\mathbf{x}, \mathbf{m}) - \tilde{\mathcal{C}}|, \delta \right)}\right),
\end{align}
where $ f(\cdot) $ could be the selection function used in previous literature, such as \eqref{Fan}--\eqref{Jiang},
$\mathcal{C}(\mathbf{x}, \mathbf{m})$ denotes the  complexity of the normalized missing region,
$\tilde{\mathcal{C}}$ is a calibration constant as shown in Fig. \ref{select} (b),
and $\delta $ is a small positive constant, chosen as 0.01 in this paper, to prevent the denominator from being zero.

Obviously, the proposed selection index does not simply rely on the loss function as before,
but is determined by the ratio of the loss function to the missingness complexity.
In this way, the low-complexity samples ignored before have a great chance to enter the training,
thus avoiding selection bias.

\begin{rmrk}
Note that $\tilde{\mathcal{C}}$ is not the mean of  the complexity distribution,
but the horizontal intercept of the dividing line of selection (see Fig. \ref{select} (b)).
It can be estimated by first applying a density clustering (e.g. DBSCAN\cite{Ester1996}) to all missingness complexity values,
and then making $\tilde{\mathcal{C}}$ equal to the minimum complexity in the largest cluster.
\end{rmrk}

After obtaining the selection criterion $S(\mathbf{x}, \mathbf{m}) $,
we can apply it in the training procedure of a deep inpainting model to help its learning.
The following algorithm illustrate this procedure when choosing $ f(\cdot ) $ as \eqref{Kawaguchi}.

\begin{algorithm}[htb]
  \caption{A pseudocode of our sample selection method }
  \label{alg}
  \begin{algorithmic}[1]
      \Require
	  A big batch size $ B$,
      a mini-batch size $ b $;	
     \For{each training iteration}
        \State Randomly choose a subset $\mathcal{B}$ of size $ B$ from the dataset;
        \For {each sample $ \mathbf{x} $ in $ \mathcal{B} $}
              \State Calculate  the corresponding selection value \eqref{select_indicate};
        \EndFor
        \State Select $ b $ samples with  the largest selection values from the subset $ B $;
        \State Update the parameters of the inpainting model with the selected samples.
    \EndFor
  \end{algorithmic}
\end{algorithm}

\begin{rmrk}
In Algorithm \ref{alg}, we use \eqref{Kawaguchi} as the selection function mainly due to its computational efficiency.
In fact, our algorithm has no special requirements on the selection function and almost all existing selection functions can be used.
\end{rmrk}
\begin{table*}[!htb]
\center
\caption{Selection bias issue of previous loss-based selection methods applied to various inpainting models. $\uparrow$ Higher is better.
$\downarrow$ Lower is better. The best results are highlighted in \textbf{bold}.}
\label{bc_t2}
\scalebox{1.0}{
\begin{tabular}{|c|ccc|ccc|ccc|}
\hline
\multicolumn{1}{|c|}{}                          &\multicolumn{3}{c|}{CelebA}                                &\multicolumn{3}{c|}{DTD}                                    &\multicolumn{3}{c|}{Paris}                                 \\
\hline
Models                                          &SSIM$\uparrow$     &PSNR$\uparrow$     &LPIPS$\downarrow$  &SSIM$\uparrow$     &PSNR$\uparrow$     &LPIPS$\downarrow$   &SSIM$\uparrow$     &PSNR$\uparrow$      &LPIPS$\downarrow$  \\
\hline
PC\cite{Liu2018A}                               &\textbf{0.8840}    &23.5596            &\textbf{0.0654}    &\textbf{0.6974}    &\textbf{21.2309}   &0.1400              &\textbf{0.7216}    &\textbf{20.2056}    &\textbf{0.1519}  \\
PC + Fan et al. \cite{Fan2017}                  &0.8587             &\textbf{23.6043}   &0.0745             &0.6282             &20.8330            &0.1434              &0.7070             &20.1457             &0.1656           \\
PC + Jiang et al. \cite{Jiang2019}              &0.8694             &23.4449            &0.0717             &0.6619             &21.1127            &\textbf{0.1350}     &0.7211             &20.1973             &0.1551           \\
PC + Kawaguchi et al. \cite{Kawaguchi2020}      &0.8793             &23.6034            &0.0674             &0.6897             &21.0353            &0.1418              &0.7176             &20.1347             &0.1521           \\
\hline
\hline
PEN-Net\cite{Zeng2019}                          &\textbf{0.8917}    &\textbf{23.3736}   &\textbf{0.0624}    &\textbf{0.7398}    &22.5731            &\textbf{0.1361}     &\textbf{0.7393}    &20.7853             &\textbf{0.1489} \\
PEN-Net + Fan et al. \cite{Fan2017}             &0.8430             &23.3659            &0.0667             &0.7076             &22.4679            &0.1411              &0.7276             &20.5467             &0.1491             \\
PEN-Net + Jiang et al. \cite{Jiang2019}         &0.8352             &23.1005            &0.0680             &0.7324             &\textbf{22.6998}   &0.1389              &0.7325             &\textbf{20.9019}    &0.1531             \\
PEN-Net + Kawaguchi et al. \cite{Kawaguchi2020} &0.8898             &23.2759            &0.0694             &0.7104             &22.5444            &0.1364              &0.7201             &20.6775             &0.1538             \\
\hline
\hline
RFR\cite{Li2020}                                &\textbf{0.9012}    &\textbf{24.1794}   &\textbf{0.0628}    &0.7355             &\textbf{23.0447}   &\textbf{0.1303}     &\textbf{0.7362}    &\textbf{21.5256}    &\textbf{0.1455}  \\
RFR + Fan et al. \cite{Fan2017}                 &0.8885             &24.1681            &0.0736             &0.7276             &22.6998            &0.1343              &0.7139             &21.5245             &0.1611           \\
RFR + Jiang et al. \cite{Jiang2019}             &0.8996             &23.6565            &0.0708             &\textbf{0.7405}    &23.0315            &0.1321              &0.7352             &21.4432             &0.1508           \\
RFR + Kawaguchi et al. \cite{Kawaguchi2020}     &0.8978             &23.7510            &0.0687             &0.7241             &22.7284            &0.1310              &0.7359             &21.5191             &0.1526           \\
\hline
\end{tabular}}
\end{table*}

\section{Experiment}

Several benchmark deep image inpainting models, including
\begin{itemize}
  \item \emph{CA} \cite{Yu2018}, which brings global features to subsequent networks through contextual attention layers.
  \item \emph{PC} \cite{Liu2018A}, which renormalizes each output to ensure that the output value is independent of the value of missing pixels in each receptive field.
  \item \emph{PEN-Net} \cite{Zeng2019}, which is a U-Net that restores images by encoding the contextual semantics of multi-scale inputs and decoding the learned semantic features back to the image.
  \item\emph{RFR} \cite{Li2020}, which is a progressive inpainting model that gradually inpaints incomplete images.
  \item \emph{RN} \cite{Yu2020}, which divides the spatial pixels into different regions according to the input mask, and adopts the mean and variance of each region for normalization.
\end{itemize}
will be used to validate the previous discussion and test the proposed selection method.
For fair comparisons, we adopt the suggest settings with existing works.

Three data sets, including CelebA \cite{Liu2018B}, DTD \cite{Cimpoi2014}, and Paris StreetView \cite{Doersch2012} datasets are considered.
Following previous research, we apply  5-cross validations on  CelebA and DTD datasets,
while  on Paris StreetView, a training set of 14900 images and a test set of 100 images are fixed.
For all training images, we resize the images to $128\times 128$ and then compress the pixels to $[0, 1]$.
If no special instructions,
all experiments are performed under a 25\% regularization mask (half the image length and width).
The complexity of missing regions are calculated by SI \eqref{sirms_eq},
EG \eqref{glcm_eq} and TV \eqref{tv_eq}, respectively.

The selection function in our algorithm is chosen as \eqref{Kawaguchi},
where the size of the random subset is set as $B=2b $,
with  $b$  denoting the batch size of  the original model.

The performance of the inpainiting results are evaluated by SSIM \cite{Wang2004},
PSNR \cite{Wang2002} and LPIPS \cite{Zhang2018}.
Here PSNR and SSIM  pay more attention to the fidelity of the image,
while LPIPS pays more attention to whether the generated image conforms to the ground truth.

\subsection{Selection bias issue}
\label{Sebias}
As discussed in Section \ref{RTM},
purely loss-based  selection methods  can not improve image inpainting models due to selection bias.
In this section,
we take three loss-based selection methods \cite{Fan2017,Jiang2019,Kawaguchi2020} as examples to illustrate this problem.
These selection methods are applied to three existing inpainting  models, respectively,
and the corresponding modification is named as  "Model +Selection method" .

It is observed from Table \ref{bc_t2} that these loss-based selection methods can hardly enhance the inpainting results,
sometimes even harmful.
Such phenomenon verifies  our judgement that solely relying on the training losses for sample selection neglects the samples of low-missingness complexity,
and can not contribute to the improvement of image inpainting models.

\subsection{Evaluation of our  proposed selection approach}

In this section, we evaluate the proposed selection approach on several benchmark image inpainting models,
only the batch selection parts of these models to be modified by Algorithm \ref{alg}.
Three image complexity metrics, including SI, EG and TV,  are used,
and the corresponding modification is named as "Model+Compelexity metric", respectively.
In order to ensure the comprehensiveness of the experiment,
both regular and irregular corrupted regions are considered.

\begin{figure}
{\label{visuala}
\includegraphics[width=1\linewidth]{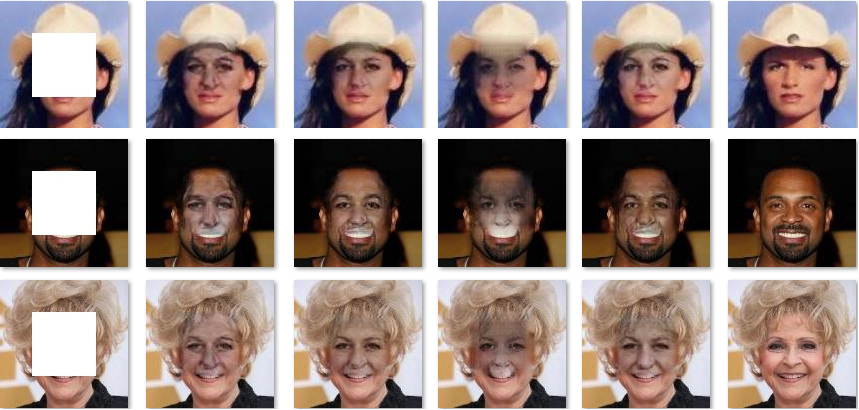}}
{\label{visualb}
\includegraphics[width=1\linewidth]{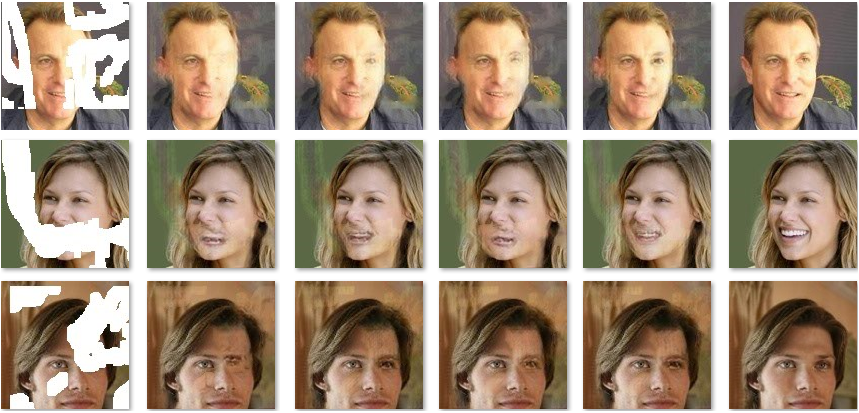}}
\subfloat{ Masked\quad PC \cite{Liu2018A} \quad PC+TV \ RFR \cite{Li2020}  RFR+TV \quad Real}
\caption{Qualitative results of the proposed selection approach using TV complexity for the inpainting models  PC \cite{Liu2018A} and RFR \cite{Li2020}  on CelebA.}
\label{visual}
\end{figure}

Fig. \ref{visual} gives an intuitive display of the inpainting results on the CelebA dataset.
It is obvious that incorporating missingness complexity can help generate better inpainted images without  "checkerboard" artifacts.

It is found from Tables \ref{bc_t11} and \ref{bc_t12} that compared with the corresponding original model,
each modified version has achieved a significant improvement under both regular and irregular masks.
This demonstrates that the proposed sample selection approach can be applied in most existing generative inpainting methods to further improve their performance.

\begin{table*}[!htb]
\center
\caption{Performance of our approach using SI, EG and TV complexity metrics for various inpainting models on \textbf{regular}  damaged images.
The best results are highlighted in \textbf{bold}.}
\label{bc_t11}
\scalebox{1.0}{
\begin{tabular}{|c|ccc|ccc|ccc|}
\hline
\multicolumn{1}{|c|}{}               &\multicolumn{3}{c|}{CelebA}                                  &\multicolumn{3}{c|}{DTD}                                       &\multicolumn{3}{c|}{Paris}                                     \\
\hline
Models                               &SSIM$\uparrow$       &PSNR$\uparrow$    &LPIPS$\downarrow$   &SSIM$\uparrow$       &PSNR$\uparrow$    &LPIPS$\downarrow$     &SSIM$\uparrow$       &PSNR$\uparrow$    &LPIPS$\downarrow$    \\
\hline
CA\cite{Yu2018}                      &0.8719               &23.5261           &0.0681              &0.7037               &21.2817           &0.1491                &0.7171               &19.8056           &0.1593               \\
CA + SI                              &0.8792               &\textbf{23.8968}  &0.0640              &0.7117               &\textbf{22.0849}  &0.1468                &0.7177               &19.9776           &0.1606               \\
CA + EG                              &0.8838               &23.3663           &0.0670              &0.7192               &21.8445           &0.1477                &0.7218               &\textbf{20.3508}  &0.1570               \\
CA + TV                              &\textbf{0.8907}      &23.7430           &\textbf{0.0617}     &\textbf{0.7221}      &21.8579           &\textbf{0.1428}       &\textbf{0.7244}      &20.1827           &\textbf{0.1541}      \\
\hline
\hline
PC\cite{Liu2018A}                    &0.8840               &23.5596           &0.0654              &0.6974               &21.2309           &0.1400                &0.7216               &20.2056           &0.1519           \\
PC + SI                              &0.8906               &23.6543           &\textbf{0.0629}     &0.6833               &21.2660           &0.1393                &0.7116               &\textbf{21.0544}  &0.1542           \\
PC + EG                              &0.8931               &23.4450           &0.0645              &0.6938               &\textbf{21.5204}  &0.1360                &0.7237               &20.9400           &0.1501           \\
PC + TV                              &\textbf{0.9101}      &\textbf{23.9411}  &0.0631              &\textbf{0.7177}      &21.4987           &\textbf{0.1350}       &\textbf{0.7373}      &20.9607           &\textbf{0.1479}  \\
\hline
\hline
PEN-Net\cite{Zeng2019}               &0.8917               &23.3736           &0.0624              &0.7398               &22.5731           &0.1361                &0.7393               &20.7853           &0.1489           \\
PEN-Net + SI                         &0.8950               &\textbf{23.8449}  &0.0633              &0.7338               &22.7281           &0.1338                &0.7374               &\textbf{21.2847}  &0.1462           \\
PEN-Net + EG                         &0.8985               &23.3750           &0.0619              &0.7404               &22.3885           &0.1345                &\textbf{0.7497}      &20.9851           &0.1471           \\
PEN-Net + TV                         &\textbf{0.9103}      &23.7959           &\textbf{0.0592}     &\textbf{0.7470}      &\textbf{22.9466}  &\textbf{0.1313}       &0.7449               &20.8579           &\textbf{0.1453}  \\
\hline
\hline
RFR\cite{Li2020}                     &0.9012               &24.1794           &0.0628              &0.7355               &23.0447           &0.1303                &0.7362               &21.5256           &0.1455           \\
RFR + SI                             &0.9046               &24.4176           &0.0639              &0.7417               &\textbf{23.3231}  &0.1294                &0.7378               &21.8334           &\textbf{0.1437}  \\
RFR + EG                             &0.8976               &24.5430           &0.0644              &0.7325               &23.0909           &0.1284                &0.7411               &\textbf{22.0250}  &0.1461           \\
RFR + TV                             &\textbf{0.9123}      &\textbf{24.7045}  &\textbf{0.0603}     &\textbf{0.7566}      &23.3045           &\textbf{0.1245}       &\textbf{0.7592}      &21.7445           &0.1443           \\
\hline
\hline
RN\cite{Yu2020}                      &0.9057               &24.4841           &0.0602              &0.7408               &23.1749           &0.1278                &0.7481               &21.5042           &0.1434           \\
RN + SI                              &0.9108               &24.7361           &0.0596              &0.7490               &24.0921           &0.1240                &0.7517               &21.7228           &0.1411           \\
RN + EG                              &0.9124               &24.5126           &0.0607              &0.7513               &24.1667           &0.1283                &0.7530               &21.3530           &0.1425           \\
RN + TV                              &\textbf{0.9271}      &\textbf{24.9621}  &\textbf{0.0571}     &\textbf{0.7627}      &\textbf{24.2567}  &\textbf{0.1222}       &\textbf{0.7689}      &\textbf{22.2263}  &\textbf{0.1387}  \\
\hline
\end{tabular}}
\end{table*}

It is worth noting that although our sample selection method brings improvements under various evaluation metrics, the magnitude of improvement is not the same.
For example, the TV-based modification seems to have a more pronounced improvements under evaluation metrics SSIM and LPIPS.

\begin{table*}[!htb]
\center
\caption{Performance of our approach using SI, EG and TV complexity metrics  for various inpainting models on \textbf{irregular} damaged images
 (with 25\% regions randomly masked by holes).
The best results are highlighted in \textbf{bold}.}
\label{bc_t12}
\scalebox{1.0}{
\begin{tabular}{|c|ccc|ccc|ccc|}
\hline
\multicolumn{1}{|c|}{}                      &\multicolumn{3}{c|}{CelebA}                                  &\multicolumn{3}{c|}{DTD}                                       &\multicolumn{3}{c|}{Paris}                       \\
\hline
Models                                     &SSIM$\uparrow$       &PSNR$\uparrow$    &LPIPS$\downarrow$   &SSIM$\uparrow$       &PSNR$\uparrow$    &LPIPS$\downarrow$     &SSIM$\uparrow$       &PSNR$\uparrow$    &LPIPS$\downarrow$  \\
\hline
CA\cite{Yu2018}                            &0.9441               &27.1526           &0.0337              &0.8119               &25.9693           &0.0710                &0.9160               &26.1059           &0.0561             \\
CA + SI                                    &0.9486               &\textbf{27.7143}  &0.0333              &0.8283               &\textbf{26.5717}  &0.0706                &0.9150               &\textbf{26.5450}  &0.0535             \\
CA + EG                                    &0.9504               &26.9176           &\textbf{0.0321}     &0.8130               &25.9124           &0.0717                &0.9245               &26.3736           &0.0524             \\
CA + TV                                    &\textbf{0.9549}      &27.4237           &0.0329              &\textbf{0.8325}      &26.5443           &\textbf{0.0649}       &\textbf{0.9332}      &26.4525           &\textbf{0.0503}    \\
\hline
\hline
PC\cite{Liu2018A}                          &0.9456               &27.1014           &0.0357              &0.8122               &26.0726           &0.0742                &0.9188               &26.0864           &0.0532           \\
PC + SI                                    &0.9464               &27.7713           &0.0339              &0.8229               &\textbf{26.7927}  &0.0734                &0.9005               &26.4712           &0.0513           \\
PC + EG                                    &0.9521               &27.6288           &0.0335              &0.8199               &25.7736           &0.0710                &0.9292               &25.9620           &0.0508           \\
PC + TV                                    &\textbf{0.9532}      &\textbf{27.8794}  &\textbf{0.0322}     &\textbf{0.8331}      &26.5176           &\textbf{0.0674}       &\textbf{0.9376}      &\textbf{26.6598}  &\textbf{0.0494}  \\
\hline
\hline
PEN-Net\cite{Zeng2019}                     &0.9461               &27.3472           &0.0346              &0.8077               &25.9995           &0.0713                &0.9177               &26.3579           &0.0547           \\
PEN-Net + SI                               &0.9499               &\textbf{28.0631}  &0.0339              &\textbf{0.8355}      &26.3568           &0.0677                &0.9239               &26.4553           &0.0550           \\
PEN-Net + EG                               &0.9518               &27.5038           &\textbf{0.0327}     &0.8226               &26.0645           &0.0692                &0.9205               &26.6886           &0.0522           \\
PEN-Net + TV                               &\textbf{0.9525}      &27.7942           &\textbf{0.0327}     &0.8341               &\textbf{26.5358}  &\textbf{0.0661}       &\textbf{0.9313}      &\textbf{26.7684}  &\textbf{0.0519}   \\
\hline
\hline
RFR\cite{Li2020}                           &0.9532               &28.0402           &0.0330              &0.8155               &26.1502           &0.0687                &0.9136               &26.2944           &0.0527           \\
RFR + SI                                   &0.9605               &28.4124           &0.0337              &0.8226               &26.3940           &0.0678                &0.9222               &26.5324           &0.0535           \\
RFR + EG                                   &0.9525               &27.7323           &0.0320              &0.8248               &26.1044           &0.0708                &0.9172               &26.4507           &0.0514           \\
RFR + TV                                   &\textbf{0.9638}      &\textbf{28.6125}  &\textbf{0.0317}     &\textbf{0.8487}      &\textbf{26.9635}  &\textbf{0.0641}       &\textbf{0.9351}      &\textbf{27.0178}  &\textbf{0.0492}  \\
\hline
\hline
RN\cite{Yu2020}                            &0.9585               &28.1042           &0.0321              &0.8258               &26.4158           &0.0670                &0.9208               &26.2332           &0.0516           \\
RN + SI                                    &0.9543               &28.4793           &0.0329              &0.8387               &\textbf{26.9671}  &\textbf{0.0628}       &0.9262               &26.4436           &0.0523           \\
RN + EG                                    &0.9583               &28.5126           &\textbf{0.0291}     &0.8283               &26.5552           &0.0673                &0.9233               &26.2195           &\textbf{0.0494}  \\
RN + TV                                    &\textbf{0.9661}      &\textbf{28.9785}  &0.0307              &\textbf{0.8461}      &26.7262           &0.0633                &\textbf{0.9349}      &\textbf{26.7281}  &0.0499           \\
\hline
\end{tabular}}
\end{table*}

\subsection{Low-missingness complexity samples are essential}

\begin{figure}[htb]
\center
\subfloat[RFR inpainting model]{\label{fig:mdleft}{\includegraphics[width=0.24\textwidth]{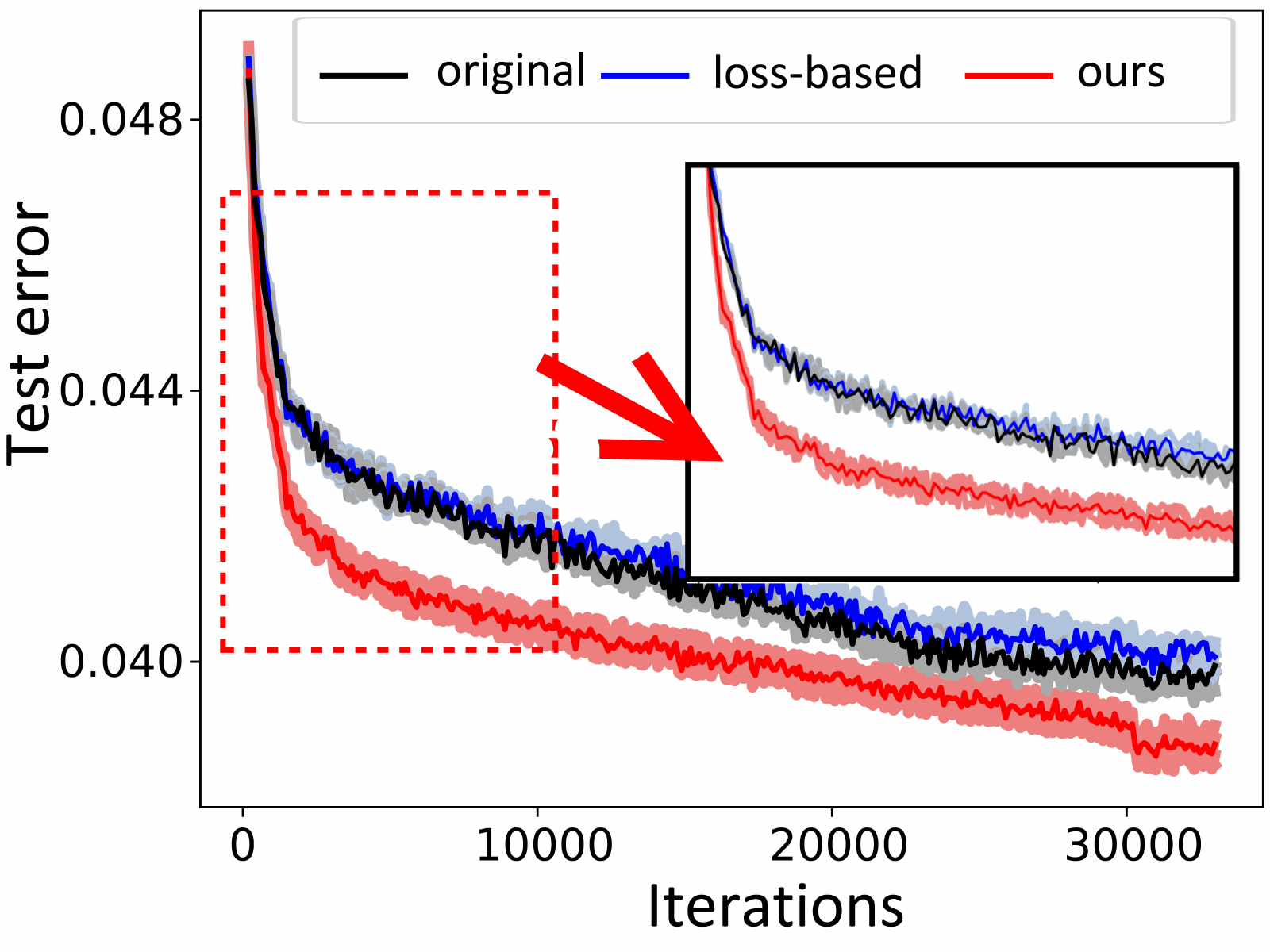}}}
\subfloat[PC  inpainting model]{\label{fig:mdleft}{\includegraphics[width=0.24\textwidth]{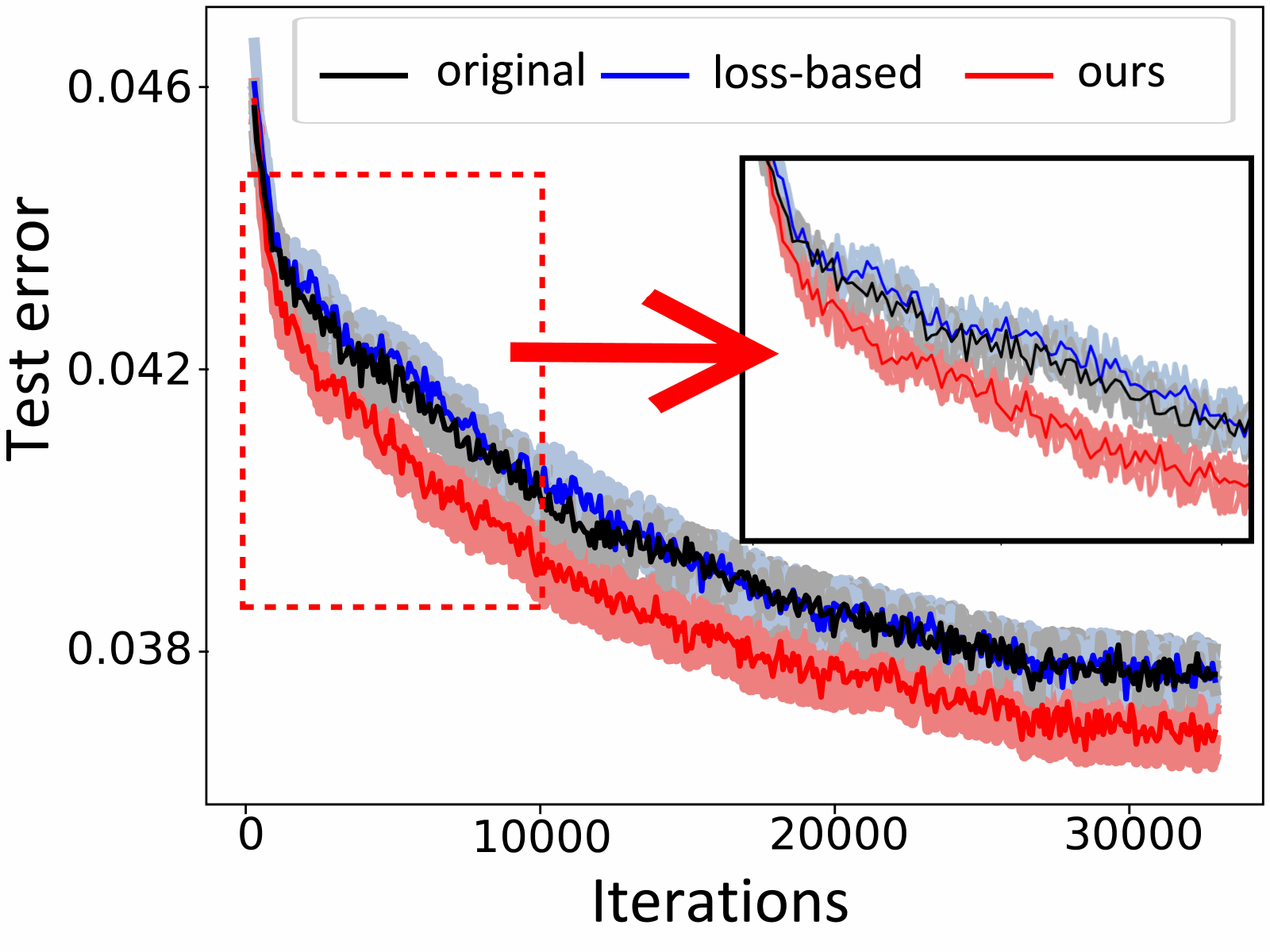}}}
\caption{Test error versus the number of iteration using the inpainting models  RFR\cite{Li2020} and PC\cite{Liu2018A} on DTD.
The lines indicate the mean values over 5 random trials,
and the shaded regions represent intervals of the sample standard deviations.}
\label{convergence_loss}
\end{figure}

We have argued that low-missingness complexity samples,
which are ignored by common loss-based selection methods,
are helpful to the learning of deep inpainting models.
To clearly verify this conclusion,
we take the loss-based selection method \cite{Kawaguchi2020} as an  example and compare the test error curves of this method with our approach based on TV complexity.
The two selection methods are adopted for  the inpainting models RFR  \cite{Li2020} and  PC \cite{Liu2018A}, respectively, on DTD dataset.

As shown in Fig. \ref{convergence_loss},
due to the use of low-missingness complexity samples,
our sample selection method expedited convergence in the early stages of model training while ultimately
achieving the lowest test error.
In contrast, the test error curve of loss-based  method is almost consistent  to that of the original model.
The above experiment results prove that  low-missingness complexity samples are vital in enhancing deep inpainting models
and should not be ignored in the training.

\subsection{Time consuming}
\label{Time}
\begin{figure}[hbt]
 \centering
 \scalebox{0.5}{\includegraphics{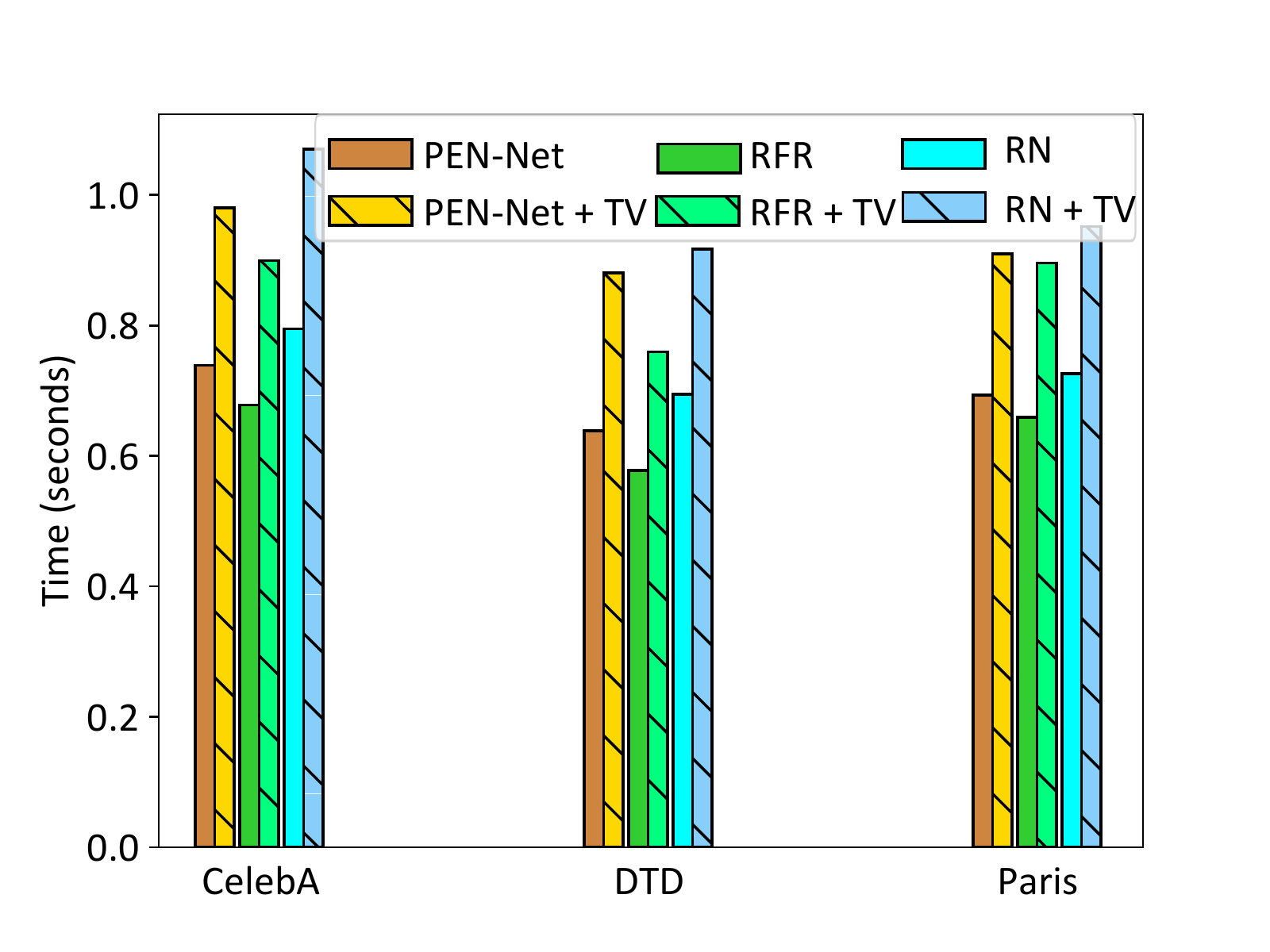}}
 \caption{Training time (seconds) per iteration using the inpainting models  PEN-Net\cite{Zeng2019}, RFR\cite{Li2020},
and RN\cite{Yu2020} on CelebA, DTD and Paris. The time is evaluated on a single NVIDIA RTX 2060s GPU.}
 \label{convergence_time}
\end{figure}

In this section, we show the computational time  of our sample selection method for different image inpainting models.
It is observed from  Fig. \ref{convergence_time} that our method is efficient and  does  not introduce too much computation cost and requires
 about 20\% extra  time for training, which mainly comes from complexity and forward loss calculations.

It is worth to noticed that Fig. \ref{convergence_time} is  a result performed on a single GPU,
but since the forward computation can be easily parallelized,
the time taken by our sample selection method can be further compressed by parallelization in practice.

\subsection{Combination of multiple complexity metrics}
Since different image complexity metrics can reflect the complexity of the image from different perspectives,
it is possible to combine multiple complexity metrics to further improve the quality of sample selection.
This subsection will examine this possibility and try a simple weighted average method.

We take the PC\cite{Liu2018A} and RFR\cite{Li2020} as image inpainting examples,
and manually coarse-tune the weights of different complexity metrics.
Since our previous experiments show that the TV  complexity usually corresponds to better results,
we set a larger weight on it.
Table \ref{bc_t3} shows that a proper combination of multiple complexity metrics may help to further improve the inpainting quality.
This finding is meaningful
which implies that if in a practical image inpainting task,
a single complexity cannot fully describe the attributes of the missing regions,
then incorporating multiple complexity measures is a feasible way.

\begin{table}[!htb]
\center
\caption{Inpainting performance using the combination of multiple complexity metrics.}
\label{bc_t3}
\scalebox{1.0}{
\begin{tabular}{|c|ccc|}
\hline
\multicolumn{1}{|c|}{}              &\multicolumn{3}{c|}{Paris}                             \\
\hline
Models                              &SSIM$\uparrow$   &PSNR$\uparrow$    &LPIPS$\downarrow$ \\
\hline
PC\cite{Liu2018A}                   &0.7216           &20.2056           &0.1519           \\
PC + SI                             &0.7116           &21.0544           &0.1542           \\
PC + EG                             &0.7237           &20.9400           &0.1501           \\
PC + TV                             &0.7373           &20.9607           &\textbf{0.1479}  \\
PC + 0.1 SI + 0.4 EG + 0.5 TV       &0.7378           &\textbf{21.1368}  &0.1484           \\
PC + 0.2 SI + 0.3 EG + 0.5 TV       &\textbf{0.7389}  &21.1141           &0.1492           \\
\hline
\hline
RFR\cite{Li2020}                    &0.7362           &21.5256           &0.1455           \\
RFR + SI                            &0.7378           &21.8334           &0.1437           \\
RFR + EG                            &0.7411           &22.0250           &0.1461           \\
RFR + TV                            &0.7592           &21.7445           &0.1443           \\
RFR + 0.2 SI + 0.3 EG + 0.5 TV      &0.7619           &\textbf{22.1274}  &\textbf{0.1421}  \\
RFR + 0.3 SI + 0.2 EG + 0.5 TV      &\textbf{0.7641}  &22.1199           &0.1430           \\
\hline
\end{tabular}}
\end{table}

The purpose of this section is mainly to demonstrate the possibility of improving performance by combining various complexities.
Although it is possible to further improve the model performance by parameterizing the weights and automatically fine-tuning them during the training,
this will bring much more computational cost,
and moreover, as can be seen from Table \ref{bc_t3},
the expected further performance improvement of the fine-tuning method is limited.

\subsection{Hyperparameter analysis}
\label{Ablation_B}
\begin{figure}[htb]
\center
\subfloat[CelebA]{\label{fig:mdleft}{\includegraphics[width=0.23\textwidth]{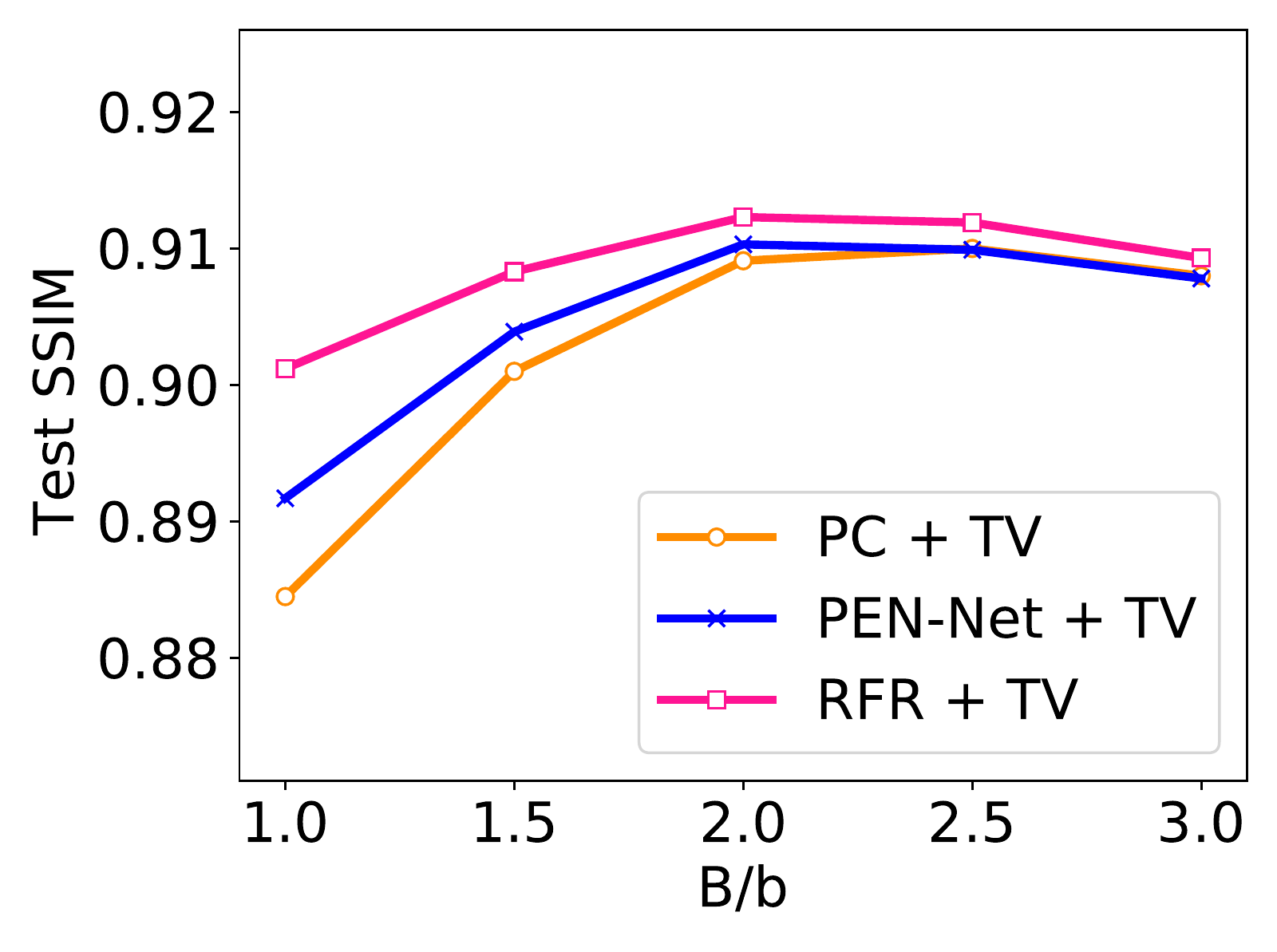}}}
\subfloat[Paris]{\label{fig:mdleft}{\includegraphics[width=0.23\textwidth]{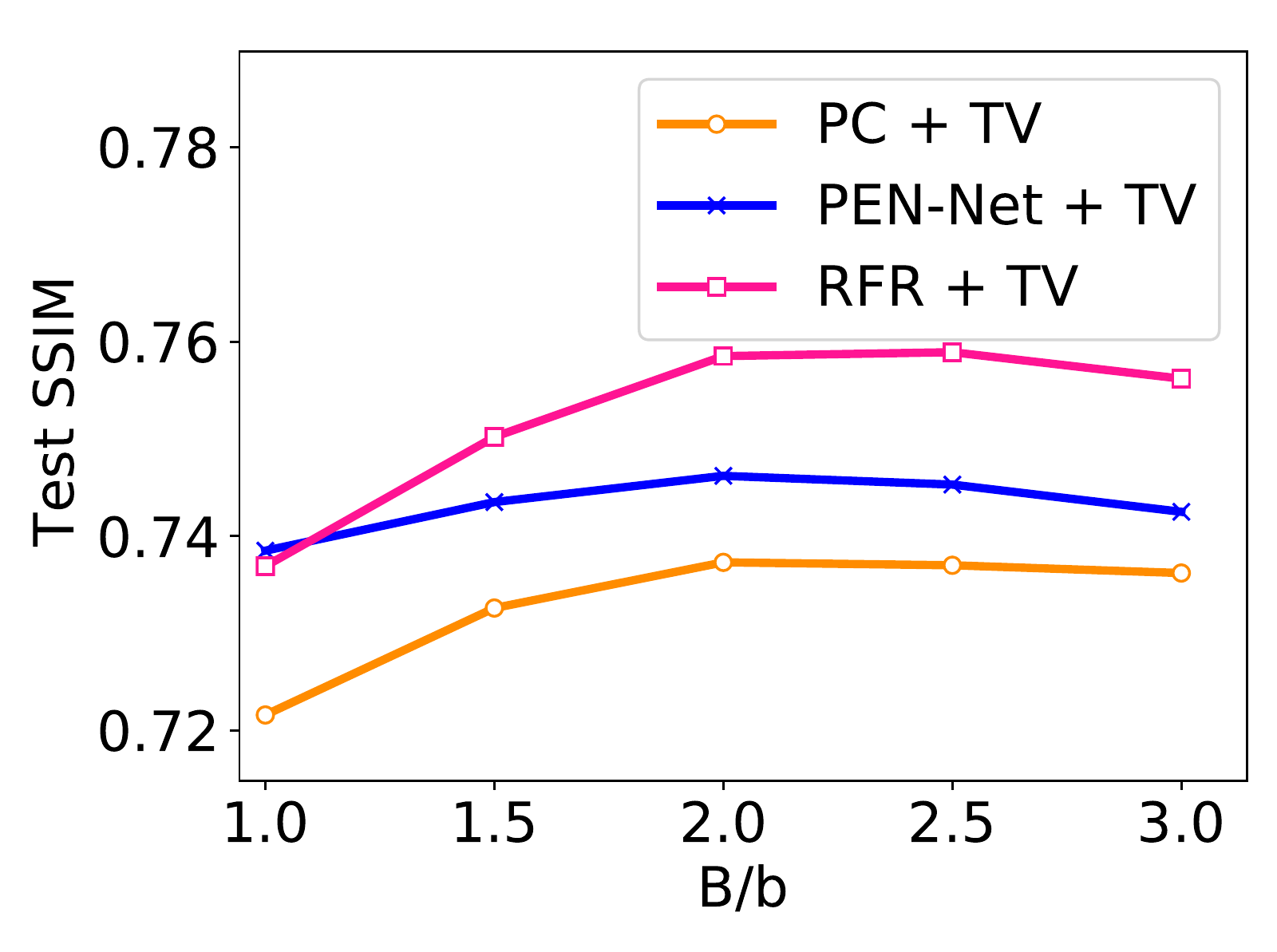}}}
\caption{Test SSIM versus the ratio of $B$ to $b$ ($B/b$) using the inpainting models  PC\cite{Liu2018A}, PEN-Net\cite{Zeng2019},
and RFR\cite{Li2020} on CelebA and Paris.}
\label{ablation_B}
\end{figure}

Our method selects $b$ training samples from a random subset of size $ B $ in each iteration.
The ratio of $B$ to $b$ can affect the inpainting performance.
In this section, we will examine the effect of this ratio.
Experimental results are summarized in Fig. \ref{ablation_B}.

It can be found that as $B$  increases from $b$ to $3b$,
the performance of the inpainting model first increases and then slowly decreases.
This implies that as the sample selection range expands, at the beginning, our approach can find more helpful samples in each iteration.
But when the size of the subset $B$ is too large,
it not only brings more computational cost,
but also makes the selection difficult and thus degrades the performance.
From Fig.\ref{ablation_B} we can empirically conclude that when $B=2b$,
the proposed method is able to balance effectiveness and efficiency.

\section{Conclusion}
In this work, we improve the quality of image inpainting from the perspective of sample selection.
We analyze the limitations of existing sample selection methods in image inpainting tasks
and propose a selection method that simultaneously fuses traditional image complexity and training loss.
Moreover, we find that the proper complexity metric is different for different data set,
and it is possible to further improve the model performance by using a combination of multiple complexity metrics.
Extensive quantitative and qualitative comparisons are conducted,
which demonstrates the superiority of the method in both performance and efficiency.
The proposed sample selection method has good interpretability  and is easy to implement,
only a few extra lines of code are needed,
and it can be used as a plug-and-play component for many deep image inpainting networks.

As a general framework illustration,
this paper only considers common complexity measures on some benchmark image datasets.
For certain practical inpainting problems such as medical images analysis,
it is suggested to adopt specific complexity measures closely related to the dataset.
Of course, if needed,
image complexity can also be obtained through learning, although this will bring additional computational cost.

Our work  shows  once again that integrating classical computer vision tools into deep learning is a promising area.
We expect it will motivate more research into performance enhancement along this direction.

\section*{Acknowledgment}
This research is partially supported by National Natural Science Foundation of China (11771257) and Natural Science Foundation of Shandong Province (ZR2018MA008)


\begin{thebibliography}{1}
\bibitem{Aouat2021}
Aouat S, Ait-hammi I, Hamouchene I. (2021).
A new approach for texture segmentation based on the gray level co-occurrence matrix.
\textit{Multimed. Tools Appl.}, 80, 24027--24052.

\bibitem{Aujol2009}
Aujol J F. (2009)
Some first-order algorithms for Total Variation based image restoration.
\textit{J. Math. Imaging Vis.}, 34(3): 307--327.


\bibitem{Ballester2001}
Ballester C, Bertalmio M, Caselles V, Sapiro G, Verdera J. (2001).
Filling-in by joint interpolation of vector fields and gray levels.
\textit{IEEE Trans. Image Process.}, 10(8): 1200--1211.


\bibitem{Bertalmio2003}
Bertalmio M, Vese L, Sapiro G, Osher S (2003).
Simultaneous structure and texture image inpainting.
In \textit{International Conference on Computer Vision and Pattern Recognition (CVPR).}


\bibitem{Chamb2012}
Chambolle A, Levine S E, Lucier B J. (2012).
An upwind finite-difference method for total variation-based image smoothing.
\textit{SIAM J. Imaging Sci.}, 4(1):277--299.


\bibitem{Chang2017}
Chang H S, Learned-Miller E, McCallum A. (2017).
Active bias: training more accurate neural networks by emphasizing high variance samples.
In \textit{Advances in Neural Information Processing Systems (NeurIPS).}



\bibitem{Chochia2015}
Chochia P A, Milukova O P. (2015).
Comparison of two-dimensional variations in the context of the digital image complexity assessment.
\textit{J. Commun. Tech. Electronics}, 60(12): 1432--1440.


\bibitem{Cimpoi2014}
Cimpoi M, Maji S, Kokkinos I, Kokkinos I, Mohamed S, Vedaldi A. (2014).
Describing textures in the wild.
In \textit{International Conference on Computer Vision and Pattern Recognition (CVPR).}

\bibitem{Ding2019}
Ding D, Ram S, Rodri\'guez J J. (2019).
Image inpainting using nonlocal texture matching and nonlinear filtering.
\textit{IEEE Trans. Image Process}, 28(4): 1705-1719.

\bibitem{Dong2013}
Dong W, Shi G, Li X. (2013).
Nonlocal image restoration with bilateral variance estimation: a low-rank approach.
\textit{IEEE Trans. Image Process.}, 22(2): 700-711.


\bibitem{Doersch2012}
Doersch C, Singh S, Gupta A, Sivic J, Efros A. (2015).
What makes paris look like paris?
\textit{Commun. ACM}, 58(12):103--110.


\bibitem{Ester1996}
Ester M, Kriegel H P, Sander J, Xu X. (1996).
Density-based spatial clustering of applications with noise.
In \textit{ACM Knowledge Discovery and Data Mining (SIGKDD)}.

\bibitem{Fan2017}
Fan Y, Lyu S, Ying Y, Hu B G. (2017).
Learning with average top-k loss.
In \textit{Advances in Neural Information Processing Systems (NeurIPS).}

\bibitem{Ghorai2019}
Ghorai M, Samanta S, Mandal S, Chanda B.(2019).
Multiple pyramids based image inpainting using local patch statistics and steering kernel feature.
\textit{IEEE Trans. Image Process}, 28(11): 5495-5509

\bibitem{Haralick1973}
Haralick R M, Shanmugam K, Dinstein I H. (1973).
Textural features for image classification.
\textit{IEEE Trans. Cybernetics}, 3(6): 610--621.

\bibitem{Huang2014}
Huang J B, Kang S B, Ahuja N, Kopf J. (2014).
Image completion using planar structure guidance.
\textit{ACM Trans. Graphic.}, 33(4): 1--10.


\bibitem{Jam2021}
Jam J, Kendrick C, Drouard V, Walker K, Hsu G S, Yap M H. (2021).
R-mnet: A perceptual adversarial network for image inpainting.
In \textit{International Conference on Computer Vision and Pattern Recognition (CVPR).}

\bibitem{Jiang2019}
Jiang A H, Wong D L K, Zhou G, et al. (2019).
Accelerating deep learning by focusing on the biggest losers.
ArXiv preprint arXiv:1910.00762.


\bibitem{Johnson2018}
Johnson T B, Guestrin C. (2018).
Training deep models faster with robust, approximate importance sampling.
In \textit{Advances in Neural Information Processing Systems (NeurIPS).}

\bibitem{Katharopoulos2018}
Katharopoulos A, Fleuret F. (2018).
Not all samples are created equal: Deep learning with importance sampling.
In \textit{International Conference on Machine Learning (ICML).}

\bibitem{Kawaguchi2020}
Kawaguchi K, Lu H. (2020).
Ordered SGD: a new stochastic optimization framework for empirical risk minimization.
In \textit{International Conference on Artificial Intelligence and Statistics (AISTATS).}

\bibitem{Kazakova2004}
Kazakova N, Margala M, Durdle N G. (2004).
Sobel edge detection processor for a real-time volume rendering system.
In \textit{IEEE International Symposium on Circuits and Systems (ISCAS).}


\bibitem{Li2017}
Li H, Luo W, Huang J. (2017).
Localization of diffusion-based inpainting in digital images.
\textit{IEEE Trans. Inf. Foren. Sec.}, 12(12): 3050-3064.

\bibitem{Li2020}
Li J, Wang N, Zhang L, Du B, Tao D. (2020).
Recurrent feature reasoning for image inpainting.
In \textit{International Conference on Computer Vision and Pattern Recognition (CVPR).}



\bibitem{Lin2017}
Lin  T Y, Goyal P, Girshick R B, He K, and Doll\'{a}r P. (2017).
Focal loss for dense object detection.
In \textit{International Conference on Computer Vision (ICCV).}


\bibitem{Liu2018A}
Liu G, Reda F A, Shih K J, Wang T C, Tao A, Catanzaro B. (2018).
Image inpainting for irregular holes using partial convolutions.
In \textit{European Conference on Computer Vision (ECCV).}


\bibitem{Liu2018B}
Liu Z, Luo P, Wang X, Tang X. (2015).
Deep learning face attributes in the wild.
In \textit{International Conference on Computer Vision (ICCV).}


\bibitem{Mindermann2021}
Mindermann S, Razzak M, Xu W, et al. (2021).
Prioritized training on points that are learnable, worth learning, and not yet learned.
ArXiv preprint arXiv:2107.02565.


\bibitem{Nazeri2019}
Nazeri K, Ng E, Joseph T, Qureshi F, Ebrahimi M. (2019).
Edgeconnect: Structure guided image inpainting using edge prediction.
In \textit{International Conference on Computer Vision Workshops}.


\bibitem{Perkio2009}
Perki J, Hyvrinen A. (2009).
Modelling image complexity by independent component analysis, with application to content-based image retrieval.
In \textit{International Conference on Artificial Neural Networks (ICANN).}


\bibitem{Pathak2016}
Pathak D, Krahenbuhl P, Donahue J, Darrell T, Efros A A. (2016).
Context encoders: Feature learning by inpainting.
In \textit{International Conference on Computer Vision and Pattern Recognition (CVPR).}


\bibitem{Romero2012}
Romero J, Machado P, Carballal A, Santosa A. (2012).
Using complexity estimates in aesthetic image classification.
\textit{J. Math. Arts}, 6(2-3): 125--136.


\bibitem{Schaul2016}
Schaul T, Quan J, Antonoglou I, Silver D. (2016).
Prioritized experience replay.
In \textit{International Conference on Learning Representations (ICLR).}


\bibitem{Shetty2018}
Shetty R, Fritz M, Schiele B. (2018).
Adversarial scene editing: Automatic object removal from weak supervision.
In \textit{Advances in Neural Information Processing Systems (NeurIPS).}

\bibitem{Shrivastava2016}
Shrivastava A, Gupta A, Girshick R. (2016).
Training regionbased object detectors with online hard example mining.
In \textit{International Conference on Computer Vision and Pattern Recognition (CVPR).}

\bibitem{Song2014}
Song Q, Chen Z, Sun S, Hu J, Cheng H. (2014).
A scene recognition method based on image complexity.
In \textit{Proceedings of SPIE--The International Society for Optical Engineering}.


\bibitem{Song2019}
Song L, Cao J, Song L, Hu Y, He R. (2019).
Geometry-aware face completion and editing.
In \textit{Conference on Artificial Intelligence (AAAI).}


\bibitem{Song2020}
Song H, Kim M, Kim S, Lee J G. (2020).
Carpe diem, seize the samples uncertain " at the moment" for adaptive batch selection.
In \textit{International Conference on Information and Knowledge Management (CIKM).}


\bibitem{Wang2002}
Wang Z, Bovik A C. (2002).
A universal image quality index.
\textit{IEEE Signal Proc. Let.}, 9(3): 81-84.


\bibitem{Wang2004}
Wang Z, Bovik A C, Sheikh H R, Simoncelli E P. (2004).
Image quality assessment: from error visibility to structural similarity.
\textit{IEEE Trans. Image Process.}, 13(4): 600-612.

\bibitem{Wang2018}
Wang C, Xu C, Wang C, Tao D. (2018).
Perceptual adversarial networks for image-to-image transformation.
\textit{IEEE Trans. Image Process.}, 27(8): 4066-4079.

\bibitem{Yan2018}
Yan Z, Li X, Li M, Zuo W, Shan S.(2018).
Shift-net: image inpainting via deep feature rearrangement.
In \textit{European Conference on Computer Vision (ECCV).}

\bibitem{Yu2018}
Yu J, Lin Z, Yang J, Shen X, Lu X, Huang T S. (2018).
Generative image inpainting with contextual attention.
In \textit{International Conference on Computer Vision and Pattern Recognition (CVPR).}

\bibitem{Yu2013}
Yu H, Winkler S. (2013).
Image complexity and spatial information.
In \textit{International Workshop on Quality of Multimedia Experience (QoMEX).}

\bibitem{Yu2020}
Yu T, Guo Z, Jin X, Wu S, Chen Z, Li W, Zhang Z, Liu S. (2020).
Region normalization for image inpainting.
In \textit{Conference on Artificial Intelligence (AAAI).}


\bibitem{Zeng2019}
Zeng Y, Fu J, Chao H, Guo B. (2019).
Learning pyramid-context encoder network for high-quality image inpainting.
In \textit{International Conference on Computer Vision and Pattern Recognition (CVPR).}

\bibitem{Zhang2018}
Zhang R, Isola P, Efros A A, et al. (2018).
The unreasonable effectiveness of deep features as a perceptual metric.
In \textit{International Conference on Computer Vision and Pattern Recognition (CVPR).}

\bibitem{Zhou2021}
Zhou Y, Barnes C, Shechtman E, Amirghodsi S. (2021).
TransFill: reference-guided image inpainting by merging multiple color and spatial transformations.
In \textit{International Conference on Computer Vision and Pattern Recognition (CVPR).}









\end{thebibliography}
\end{document}